\documentclass[sigconf, nonacm]{acmart}

\usepackage{listings}
\usepackage{graphicx}
\usepackage{placeins}
\usepackage{natbib}
\usepackage{rotating}
\usepackage{lscape}
\usepackage{subcaption} % for setting the space between multiple figures

\usepackage{pdflscape}
\usepackage{afterpage}
\usepackage{capt-of}% or use the larger `caption` package
\usepackage{tabularx}
\usepackage{tablefootnote}
\usepackage{multirow}
\usepackage{appendix}
\usepackage{multicol} % Load the 'multicol' package
\usepackage{changepage} % Load the 'changepage' package

\begin{document}

\title{UnScientify: Detecting Scientific Uncertainty in Scholarly Full Text}

%%
%% The "author" command and its associated commands are used to define the authors and their affiliations.
\author{Panggih Kusuma Ningrum}\authornote{Corresponding author}
\orcid{0000-0002-8630-6603}
\affiliation{%
  \institution{Université de Franche-Comté, CRIT}
  \city{F-25000 Besancon}
  \country{France}
}
\email{panggih_kusuma.ningrum@univ-fcomte.fr}

\author{Philipp Mayr}
\orcid{0000-0002-6656-1658}
\affiliation{%
  \institution{GESIS –- Leibniz Institute for the Social Sciences}
  \city{Cologne}
  \country{Germany}
}
\email{philipp.mayr@gesis.org}

\author{Iana Atanassova}
\orcid{0000-0003-3571-4006}
\affiliation{%
  \institution{Université de Franche-Comté, CRIT}
  \city{F-25000 Besancon}
  \country{France}
}
\affiliation{%
  \institution{Institut Universitaire de France (IUF)}
    \country{France}
}
\email{iana.atanassova@univ-fcomte.fr}

\begin{abstract}

This demo paper presents UnScientify \footnote{Demo app: \url{https://bit.ly/unscientify-demo}}, an interactive system designed to detect scientific uncertainty in scholarly full text. The system utilizes a weakly supervised technique that employs a fine-grained annotation scheme to identify verbally formulated uncertainty at the sentence level in scientific texts. The pipeline for the system includes a combination of pattern matching, complex sentence checking, and authorial reference checking. Our approach automates labeling and annotation tasks for scientific uncertainty identification, taking into account different types of scientific uncertainty, that can serve various applications such as information retrieval, text mining, and scholarly document processing.
Additionally, UnScientify provides interpretable results, aiding in the comprehension of identified instances of scientific uncertainty in text.

\end{abstract}


\maketitle

\section{Introduction}

Uncertainty is an inherent part of scientific research, as the very nature of scientific inquiry involves posing questions, developing hypotheses, and testing them using empirical evidence. Despite the best efforts of scientists to control for extraneous variables and obtain accurate measurements, there is always a certain degree of uncertainty associated with any scientific findings. This uncertainty can arise from a variety of sources, such as measurement error, sampling bias, or limitations in experimental design. Consequently, researchers resort to various strategies to manage and mitigate uncertainty when presenting their findings in academic articles. These may include using language that is overly definitive or hedging their claims with qualifiers such as "presumably" or "possible" \citep{hyland_talking_1996}.

The identification of Scientific Uncertainty (SU) in scientific text is a crucial task that can provide insights into the reliability and validity of scientific claims, help in making informed decisions, and identify areas for further investigation. Besides, detecting uncertainty has become a significant aspect of the peer-review process, which serves as a gatekeeper for the dissemination of scientific knowledge. However, the identification of scientific uncertainty in text is a complex task that requires expertise in linguistics and scientific knowledge, and is often time-consuming and labor-intensive. The primary issue stems from the fact that handling unstructured textual data in scientific literature is complicated. Previous research has mainly focused on identifying a specific set of uncertainty cues and markers in scientific articles, using a particular section of the text, such as the abstract \citep{vincze_bioscope_2008} or the full text \citep{medlock_weakly_nodate,riccioni_self-mention_2021}. These studies have helped expand the vocabulary and lexicon associated with uncertainty. However, their practical application is often inaccurate because of the intricate nature of natural language.

More sophisticated automation techniques such as machine learning and deep learning have undoubted potential for dealing with Natural Language Processing (NLP) tasks. However, the task of scientific uncertainty identification is challenging due to several factors. Firstly, there is a scarcity of available extensively annotated corpus that can be used by such techniques for scientific uncertainty identification. At present, certain corpora are limited in their scope as they only capture a particular type of uncertainty within a specific domain. For example, the BioScope corpus concentrates solely on negation or uncertainty in biological scientific abstracts \citep{vincze_bioscope_2008}, while the FACTBANK corpus is designed to identify the veracity or factuality of event mentions in text \citep{sauri_factbank_2009}. Similarly, the Genia Event corpus is restricted to the annotation of biological events with negation \citep{kim_corpus_2008}. Therefore, there is a need for more diverse corpora that capture a wider range of uncertainty types and domains, to facilitate a more comprehensive understanding of uncertainty in natural language processing. 

Secondly, identifying scientific uncertainty in text involves complex linguistic features as it is often conveyed through a combination of linguistic cues, including the use of modal verbs (e.g. may, could, might), hedging devices (e.g. seems, appears, suggests), and epistemic adverbs (e.g. possibly, probably, perhaps) \citep{chen_scalable_2018, bongelli_writers_2019}. Identifying such linguistic markers of uncertainty is not always straightforward, as they can be expressed in a variety of ways depending on the writing style or stance of the scientist.

Another challenge concerns scientists' discourse in scientific writing. A typical scientific text contains various statements and information which not only discuss the current or present study but also the former studies \citep{stocking_constructing_1993}. While writing the article, scientists can use uncertainty claims from other studies as a rhetorical tool to persuade others or to describe and organize some state of knowledge. As a result, distinguishing the reference of the uncertainty feature -- whether the statement actually demonstrates uncertainty in the current study or in the former study, is a crucial factor in better understanding the context of scientific uncertainty. A study conducted by \citet{bongelli_writers_2019} is one of few that was aware of this concern. In more detail, this study only focused on the certainty and uncertainty expressed by the speakers/writers in the here-and-now of communication and excluded those that were expressed by the other party.

To overcome these challenges, we propose a weakly supervised technique that employs a fine-grained annotation scheme to construct a system for scientific uncertainty identification from scientific text focusing on the sentence level. Our approach can be used to automate labeling or annotating tasks for scientific uncertainty identification. Moreover, our annotation scheme provides interpretable results, which can aid in the understanding of the identified instances of scientific uncertainty in text. We anticipate that our approach will contribute to the development of more accurate and efficient scientific uncertainty identification systems, and facilitate the analysis and interpretation of scholarly documents in NLP.

\section{Data}

The present study employs three annotated corpora as the training set. These corpora consist of 59 journals from four different disciplines: Medicine, Biochemistry, Genetics \& Molecular Biology, Multidisciplinary, and Empirical Social Science\footnote{All social science articles are from SSOAR (\url{https://www.ssoar.info/}); we selected articles from 53 social science journals indexed in SSOAR.} which represent Science, Technology, and Medicine (STM) as well as Social Sciences and Humanities (SSH). The corpora consist of 1001 randomly selected English sentences from 312 articles across 59 journals. These sentences were annotated to identify uncertainty expressions and authorial references. By utilizing multiple corpora from different disciplines, this study aims to capture a diverse range of uncertainty expressions and improve the generalizability of the results. Table \ref{table:corpora} illustrates the distribution of the data in the corpora and Table \ref{table:sample_sentences} shows the sample of annotated sentences.

\begin{table*}[htbp]
    \centering
   % \begin{tabularx}{\columnwidth}{p{1.9cm}p{1.8cm}rr}
    \begin{tabularx}{\textwidth}{XXrr}
        \toprule
        Discipline & Journal &Articles&Sentences\\
        \midrule
        Medicine & BMC Med & 51 & 95 \\
         & Cell Mol Gastroenterol Hepatol & 25 & 36 \\
        Biochemistry, Genetics \& Molecular Biology & Nucleic Acids Res & 52 & 63 \\
         & Cell Rep Med & 22 & 48 \\
        Multidisciplinary & Nature & 34 & 57 \\
         & PLoS One & 42 & 55 \\
        Empirical Social Science & SSOAR (53 journals) & 86 & 647 \\
        \bottomrule
    \end{tabularx}
    \caption{Corpora description}
    \label{table:corpora}
\end{table*}

\begin{table*}[htbp]
    \centering
   % \begin{tabularx}{\columnwidth}{p{4.5cm}p{0.6cm}p{0.8cm}}
     \begin{tabularx}{\textwidth}{Xll}
        \toprule
        Sentence & SU Check & Authorial Ref. \\
        \midrule
        It is possible that corticosteroids prevent some acute gastrointestinal complications. & Yes & Author(s) \\
        However, we find no evidence to support this hypothesis either. & No & - \\
        But, how this kind of coverage might influence the "we" feeling among Europeans, still remains somehow an open question. & Yes & Author(s) \\
        Previous meta-analyses have shown a significant benefit for NaHCO3 in comparison to normal saline (NS) infusion [6,7], although they highlighted the possibility of publication bias. & Yes & Former/Prev. Study(s) \\
        \bottomrule
    \end{tabularx}
    \caption{Samples of annotated sentences}
    \label{table:sample_sentences}
\end{table*}

\section{Approach}

Identifying scientific uncertainty in academic texts is a complex task due to various reasons. Previous research indicates that relying solely on cues or markers such as hedging words or modal verbs may not accurately identify scientific uncertainty \citep{Ningrum2023}. The natural language and writing styles used by scientists, along with variations in domain-specific terminology, add to the complexity of identifying uncertainty in scientific text. Moreover, the lack of clear boundaries for expressions of uncertainty makes n-gram-based approaches too inflexible to capture the various forms and expressions of uncertainty in scientific language. To address these limitations, our research proposes a fine-grained annotation scheme for identifying uncertainty in scientific texts. %This scheme aims to provide a more accurate and comprehensive approach to identifying uncertainty in scientific language.

\subsection{Fine-grained SU annotation scheme and patterns formulation}

The present study adopts a span-based approach for identifying scientific uncertainty in academic text. Rather than relying solely on linguistic cues, the scheme classifies spans of text into several groups based on their linguistic features, including Part of Speech (POS) tags, morphology, and dependency. The scheme is also informed by a comprehensive analysis of scientific language, allowing for a more nuanced and accurate understanding of uncertainty expression. 

\begin{figure}[!hbt]
  \centering
  \includegraphics[width=0.9\columnwidth]{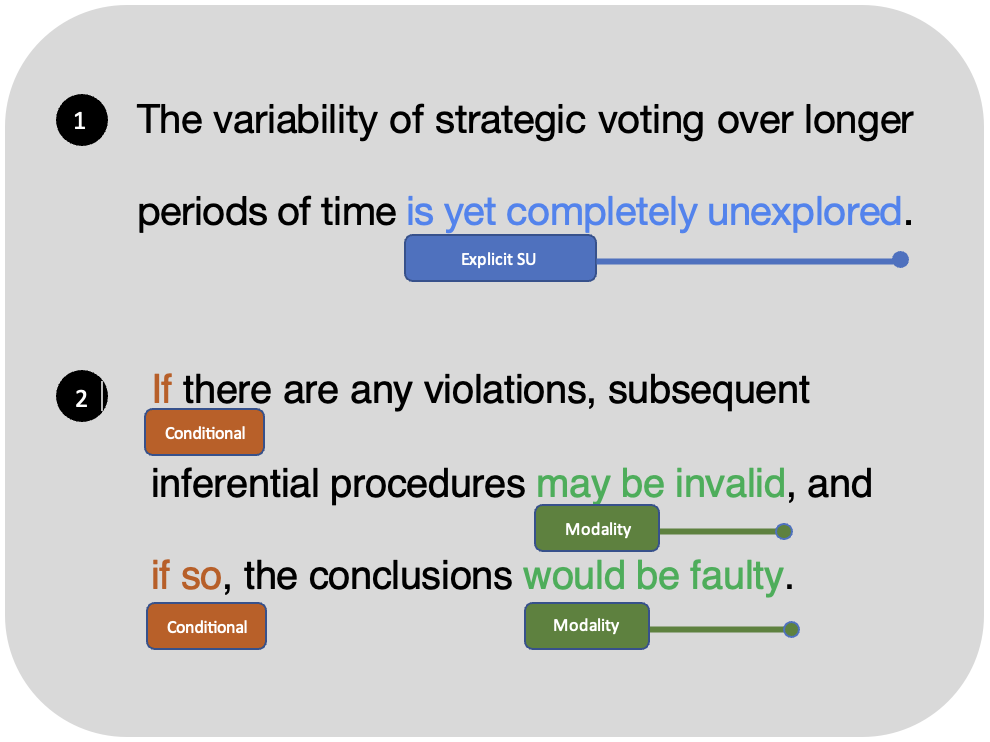}
  \caption{Two annotated sentences with SU expressions. Samples of output from span annotation process are shown in different colours based on their SU Pattern Group.}
  \label{fig:Span_labelling}
\end{figure}

During the annotation process, a list of annotated spans was created and classified into twelve groups of scientific uncertainty (SU) patterns based on their semantic meaning and characteristics. The groups include conditional expressions, hypotheses, predictions, and subjectivity, among others. In other words, the classification is based on the types of expressions used to convey uncertainty and the context in which they are used. Additionally, the scheme considers spans of text that signal disagreement statements as one of SU groups, despite ongoing debate regarding whether disagreement expressions should be considered as such. The justification for this approach is rooted in the idea that uncertainty in research can stem from conflicting information or data, where multiple sources provide contradictory knowledge \citep{zimmermann_application-oriented_2000}. This type of uncertainty cannot be reduced by increasing the amount of information. Once the annotated spans are classified, Scientific Uncertainty Span Patterns (SUSP) are formulated based on the word patterns of each span and its linguistic features. Figure \ref{fig:Span_labelling} illustrates the output from the spans annotation process.

Figure \ref{fig:Span_labelling} shows the application of span annotation to identify scientific uncertainty in each sentence. Each span is assigned a label corresponding to its SU pattern group. It should be noted that a sentence can have multiple labels assigned to different SU pattern groups, as seen in the second example, where labels for both conditional expression and modality are present. This feature of our annotation scheme enables the identification of complex expressions of uncertainty in scientific text. Table \ref{tab-examples} shows more details about the list of SU pattern groups and samples from each group and more detailed information about the pattern formulation process can be seen in the demo's documentation \footnote{Demo's documentation: \url{https://bit.ly/unscientify-demo}}.

\begin{table*}[htbp]
     \begin{tabularx}{\textwidth}{>{\hsize=0.5cm}llXX}
            \toprule
            No & Pattern Group & Description & Examples \\ 
            \midrule
            1 & Explicit SU & \multirow{2}{6.5cm}{Explicit SU group displays expressions with obvious scientific uncertainty keywords, indicating direct and explicit uncertainty expression} & 1) In addition, the role of the public \textbf{is often unclear}. \\
              & & & 2) ... the functional relevance of G4 in vivo in mammalian cells \textbf{remains controversial}.\\ 
            2 & Modality & \multirow{2}{6.5cm}{The modality group comprises expressions that indicate uncertainty through the use of modal language} & 1) Different voters \textbf{might have} different interpretations about ... \\
              & & & 2) There \textbf{may also be} behavioral effects. \\ 
            3 & \multirow{2}{2cm}{Conditional Expression} & \multirow{2}{6.5cm}{The conditional expression group includes expressions that indicate uncertainty by presenting a condition or circumstance that must be met for a certain outcome to occur} & 1) \textbf{If} persons perceive the media as hostile, \textbf{it is probable that} the mere-exposure effect is weakened thus we hypothesize... \\
              & & & 2) \textbf{If} there are any violations, subsequent inferential procedures may be invalid, and \textbf{if so}, the conclusions would be faulty. \\
            4 & Hypothesis & \multirow{2}{6.5cm}{The hypothesis group encompasses expressions that indicate uncertainty by proposing a tentative explanation or assumption that requires further testing and verification to be confirmed} & 1) \textbf{Hypotheses} predict that aggregate support for markets should be stronger... \\
              & & & 2) \textbf{We assume} that post-materialistic individuals may have differing attitudes towards doctors than those... \\
            5 & Prediction & \multirow{2}{6.5cm}{The prediction group comprises expressions that indicate uncertainty by proposing a forecast or projection that may or may not come to fruition, thereby introducing an element of uncertainty} & 1) In July 2017, the National Grid's Future Energy Scenarios \textbf{projected that} the UK government... \\
              & & & 2) Since aging leads to decreased Sir2, we \textbf{predicted that}, in young cells... \\
            6 & \multirow{2}{2cm}{Interrogative Expression} & \multirow{2}{6.5cm}{The interrogative expression group includes expressions that indicate uncertainty by posing a question or series of questions, which may suggest doubt or uncertainty about a particular concept or phenomenon} & 1) The study aims to determine \textbf{whether} the observed results can be replicated across different populations. \\
              & & & 2) ...this research literature has also contested \textbf{whether or not} citizens' knowledge about these issues is accurate... \\ 
            7 & \multirow{2}{2cm}{Non-generalizable statement} & \multirow{2}{6.5cm}{The non-generalizable statement group expresses uncertainty with limited scope or applicability, which may not represent a broader context or population} & 1) Our study ... thus \textbf{cannot be directly generalized} to low-income nations nor extrapolated into the long-term future. \\
              & & & 2) ...estimates \textbf{may not be generalisable} to women in other to women in other ancestry groups... \\ 
            8 & Adverbial SU & \multirow{2}{6.5cm}{The scientific uncertainty group includes adverbs that modify or shift the sentence's meaning, introducing uncertainty} & 1) ...direct and indirect readout during the transition from search to recognition mode is \textbf{poorly} understood. \\
              & & & 2) It will be \textbf{quite} certain that they belong to the subpopulation of gender heterogenous... \\ 
            9 & Negation & \multirow{2}{6.5cm}{The negation group comprises expressions that indicate uncertainty through the use of negation which may alter the meaning of the sentence and introduce an element of uncertainty} & 1) The identity of C34 modification in... is \textbf{not clear}. \\
              & & & 2) There was \textbf{no consistent} evidence for a causal relationship between age at menarche and lifetime number of sexual partners... \\
            10 & Subjectivity & \multirow{2}{6.5cm}{The subjectivity group includes expressions indicating uncertainty through subjective language like opinions, beliefs, or personal experiences} & 1) \textbf{We believe that} there are good reasons for voters to care about... \\
              & & & 2) \textbf{To our knowledge}, this is the first study to provide global... \\
            11 & Conjectural & \multirow{2}{6.5cm}{The conjectural group expresses uncertainty through conjecture or speculation, using guessing or suppositions without concrete evidence} & 1) This belief \textbf{seems to be} typical for moderate religiosity. \\
              & & & 2) Better performance \textbf{seems to be linked} to life satisfaction... \\ 
            12 & Disagreement & \multirow{2}{6.5cm}{The disagreement group includes expressions that express uncertainty through disagreement or contradiction, often indicating opposing viewpoints or conflicting evidence} & 1) \textbf{In contrast to previous studies}, our results did not show a significant effect...\\
              & & & 2) \textbf{On the one hand}, some researchers argue that the use of technology in the classroom can enhance... \\
            \bottomrule
        \end{tabularx}
        \captionof{table}{SU Pattern Groups and examples of annotated sentences with SU spans written in bold}
        \label{tab-examples}
\end{table*}

\subsection{Authorial Reference Checking}

Authorial reference is crucial in scientific writing to provide context, especially when identifying scientific uncertainty. It helps to indicate the authorship of the argument and distinguish between the claims of the author and those of others. This can be achieved through various styles of authorial reference, such as in-text citations, reference or co-reference \citep{powley_evidence-based_2007}. Additionally, there are disciplinary variations in both the frequency and use of personal and impersonal authorial references \citep{khedri_how_2020}.

Proper attribution of uncertain claims is important to determine their origin and evaluate the credibility of the argument. For instance, when stating a hypothesis, it is essential to indicate whether it is the author's hypothesis or cited from another source. This helps the reader to assess the level of uncertainty associated with the hypothesis. %Overall, authorial reference is vital in scientific writing to provide more context while identifying SU and to assist readers in evaluating the credibility of the argument presented.

In the present study, the authorial reference of each sentence was annotated based on the citation \& co-citation patterns, and the use of personal \& impersonal authorial references. Furthermore, sentences were labeled into three groups including 1) author(s) of the present article, or 2) author(s) of previous research. The last group, 3) both, is intended to accommodate complex sentences that may refer to both the author(s) and the previous study(s). Here, we present some examples of typical authorial reference mentions in context:
\begin{quote}
    1. <I/We/Our study...> <text> \\
    2. <Author/The former study...> <text> \\
    3. (Author) (Year) <Text> \\
    4. <Text> (Author1, Year1; Author2, Year2 . . .) \\
    5. <Text> [Ref-No1, Ref-No2 . . . ]
\end{quote}

\section{Demo System}
%% Classifier Framework for SU detection

The demo system\footnote{The demo is publicly available on \url{https://bit.ly/unscientify-demo}.} for identifying SU expressions operates at the sentence level and consists of three main components: 1) Pattern Matching, 2) Complex Sentence Checking, and 3) Authorial Reference Checking, as shown in Figure 2.

\begin{figure*}[htbp]
    \centering
    \includegraphics[width=\textwidth]{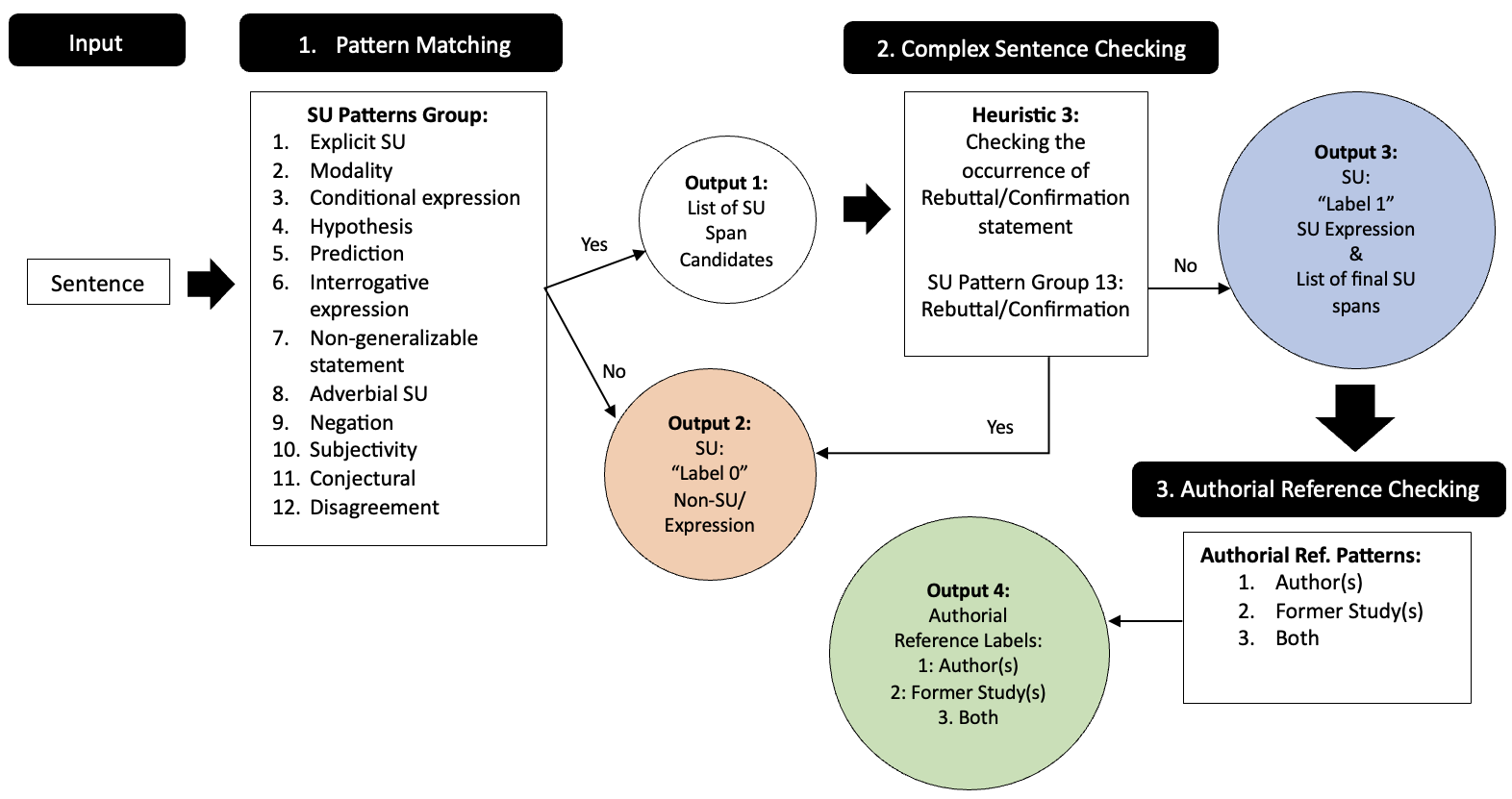}
    \caption{Scientific Uncertainty (SU) expression identification workflow}
    \label{fig:SU_iden_flow}
\end{figure*}

The first step, Pattern Matching, employs a list of patterns derived from 12 SU pattern groups (see Table \ref{tab-examples}). The input sentence is matched against these patterns, and if a match is found, a list of SU span candidates is generated. If there is no match, the sentence is labeled as 'Non-SU expression'. To optimize the matching process, we customized a rule-based matcher from Spacy, which considers both  keyword matches and patterns and linguistic features.

The second step, Complex Sentence Checking, determines whether there are any rebuttal or confirmation statements that can cancel the uncertainty expressed in the sentence. If no such statements are detected, the system labels the sentence as 'SU Expression' and provides a list of final SU spans that provide information on the reason why a particular sentence is considered a 'SU expression'.

The third step, Authorial Reference Checking, identifies the authorship of the uncertainty expression, whether it belongs to the authors, to a previous study, or both. The output of this step is the authorial reference of the sentence.

\begin{figure}[!hbt]
  \centering
  \includegraphics[width=1\columnwidth]{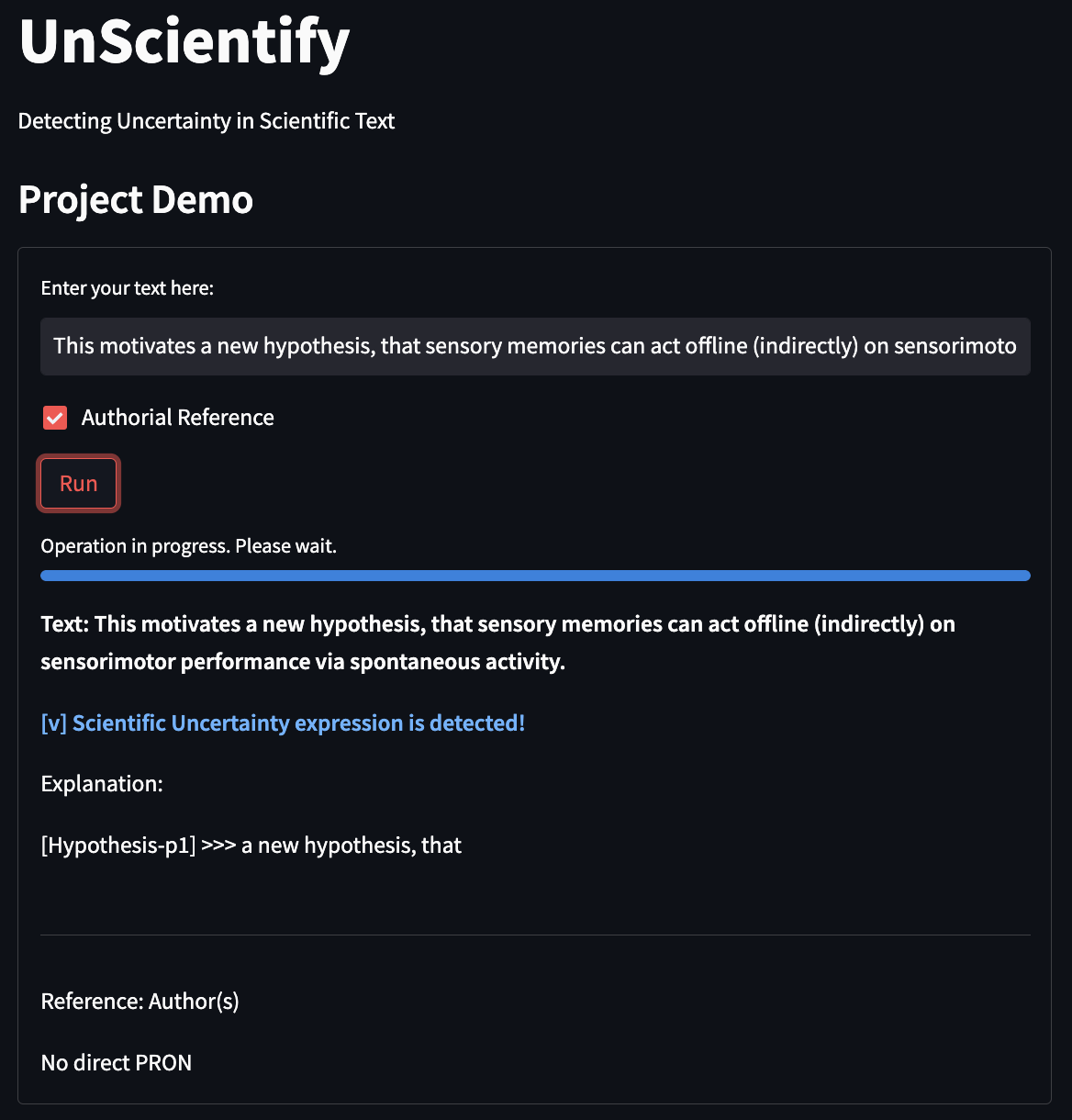}
  \caption{UnScientify demo interface with a sample sentence and annotation output}
  \label{fig:demo}
\end{figure}

Figure \ref{fig:demo} provides an overview of the functioning of UnScientify. The input sentence is annotated as an SU expression, matching the 'Hypothesis group' pattern. This demonstrates that UnScientify not only detects uncertainty expressions in sentences but also provides information about which sentence elements support the outcome as well as descriptive information about why the sentence is considered an SU expression. In this case, the output identifies the sentence as an SU expression due to the occurrence of the "Hypothesis group' pattern in the sentence, indicating a tentative explanation or assumption that requires further testing for confirmation. Additionally, UnScientify checks for authorial references, labeling this instance as 'Author(s)', suggesting that the sentence originates from the author rather than being cited from other sources or previous studies. As a result, it provides more contextual and interpretable results. Further demonstrations of UnScientify can be viewed in Appendix \ref{Appendix:unscientify_demo}.

\section{Conclusion}

Our demonstration system offers a comprehensive approach to identifying uncertainty expressions in scientific text. By utilizing pattern matching, complex sentence checking, and authorial reference checking, we provide clear and interpretable output that explains why a sentence is flagged as expressing uncertainty, addresses the element of SU expression, and verifies authorship reference.

We firmly believe that our approach holds great potential for enhancing information retrieval, text mining, and broader scientific article processing. Moreover, it lays the groundwork for further research on scientific uncertainty and epistemology. While our system currently operates at the sentence level, it can be expanded to process text at the document level.

To further enhance the UnScientify system, we acknowledge the need for improvements to identify additional dimensions of scientific uncertainty, including its nature, context, timeline, and communication characteristics. Nonetheless, we are confident that our scheme serves as a promising starting point for an in-depth exploration of how scientific knowledge is constructed and communicated.

\begin{acks}

This research was funded by the French ANR InSciM Project (2021-2024) under grant number ANR-21-CE38-0003-01, and the Chrysalide Mobilité Internationale des Doctorants (MID) mobility grant from the University of Bourgogne Franche-Comté, France. 
Our appreciation extends to the GESIS -- Leibniz Institute for the Social Sciences for providing the dataset and invaluable assistance, and to Nina Smirnova for her unwavering support throughout this project.

\end{acks}

%\clearpage

\bibliographystyle{unsrtnat}
\bibliography{sample}

\begin{thebibliography}{13}
\providecommand{\natexlab}[1]{#1}
\providecommand{\url}[1]{\texttt{#1}}
\expandafter\ifx\csname urlstyle\endcsname\relax
  \providecommand{\doi}[1]{doi: #1}\else
  \providecommand{\doi}{doi: \begingroup \urlstyle{rm}\Url}\fi

\bibitem[Hyland(1996)]{hyland_talking_1996}
Ken Hyland.
\newblock Talking to the {Academy}: {Forms} of {Hedging} in {Science}
  {Research} {Articles}.
\newblock \emph{Written Communication}, 13\penalty0 (2):\penalty0 251--281,
  April 1996.
\newblock ISSN 0741-0883.
\newblock \doi{10.1177/0741088396013002004}.
\newblock URL \url{https://doi.org/10.1177/0741088396013002004}.
\newblock Publisher: SAGE Publications Inc.

\bibitem[Vincze et~al.(2008)Vincze, Szarvas, Farkas, Móra, and
  Csirik]{vincze_bioscope_2008}
Veronika Vincze, György Szarvas, Richárd Farkas, György Móra, and János
  Csirik.
\newblock The {BioScope} corpus: biomedical texts annotated for uncertainty,
  negation and their scopes.
\newblock \emph{BMC Bioinformatics}, 9\penalty0 (S11):\penalty0 S9, December
  2008.
\newblock ISSN 1471-2105.
\newblock \doi{10.1186/1471-2105-9-S11-S9}.
\newblock URL
  \url{https://bmcbioinformatics.biomedcentral.com/articles/10.1186/1471-2105-9-S11-S9}.

\bibitem[Medlock and Briscoe(2007)]{medlock_weakly_nodate}
Ben Medlock and Ted Briscoe.
\newblock Weakly supervised learning for hedge classification in scientific
  literature.
\newblock In \emph{Proceedings of the 45th Annual Meeting of the Association of
  Computational Linguistics}, pages 992--999, Prague, Czech Republic, June
  2007. Association for Computational Linguistics.
\newblock URL \url{https://aclanthology.org/P07-1125}.

\bibitem[Riccioni et~al.(2021)Riccioni, Bongelli, and
  Zuczkowski]{riccioni_self-mention_2021}
Ilaria Riccioni, Ramona Bongelli, and Andrzej Zuczkowski.
\newblock Self-mention and uncertain communication in the {British} {Medical}
  {Journal} (1840-2007): {The} decrease of subjectivity uncertainty markers.
\newblock \emph{Open Linguistics}, 7\penalty0 (1):\penalty0 739--759, January
  2021.
\newblock ISSN 23009969.
\newblock \doi{10.1515/OPLI-2020-0179/MACHINEREADABLECITATION/RIS}.
\newblock URL
  \url{https://www.degruyter.com/document/doi/10.1515/opli-2020-0179/html?lang=en}.
\newblock Publisher: Walter de Gruyter GmbH.

\bibitem[Saurí and Pustejovsky(2009)]{sauri_factbank_2009}
Roser Saurí and James Pustejovsky.
\newblock Factbank: {A} corpus annotated with event factuality.
\newblock \emph{Language Resources and Evaluation}, 43\penalty0 (3):\penalty0
  227--268, September 2009.
\newblock ISSN 1574020X.
\newblock \doi{10.1007/s10579-009-9089-9}.

\bibitem[Kim et~al.(2008)Kim, Ohta, and Tsujii]{kim_corpus_2008}
Jin-Dong Kim, Tomoko Ohta, and Jun'ichi Tsujii.
\newblock Corpus annotation for mining biomedical events from literature.
\newblock \emph{BMC Bioinformatics}, 9\penalty0 (1):\penalty0 10, January 2008.
\newblock ISSN 1471-2105.
\newblock \doi{10.1186/1471-2105-9-10}.
\newblock URL \url{https://doi.org/10.1186/1471-2105-9-10}.

\bibitem[Chen et~al.(2018)Chen, Song, and Heo]{chen_scalable_2018}
Chaomei Chen, Min Song, and Go~Eun Heo.
\newblock A scalable and adaptive method for finding semantically equivalent
  cue words of uncertainty.
\newblock \emph{Journal of Informetrics}, 12\penalty0 (1):\penalty0 158--180,
  February 2018.
\newblock ISSN 17511577.
\newblock \doi{10.1016/j.joi.2017.12.004}.
\newblock URL
  \url{https://linkinghub.elsevier.com/retrieve/pii/S1751157717301712}.

\bibitem[Bongelli et~al.(2019)Bongelli, Riccioni, Burro, and
  Zuczkowski]{bongelli_writers_2019}
Ramona Bongelli, Ilaria Riccioni, Roberto Burro, and Andrzej Zuczkowski.
\newblock Writers’ uncertainty in scientific and popular biomedical articles.
  {A} comparative analysis of the {British} {Medical} {Journal} and {Discover}
  {Magazine}.
\newblock \emph{PLoS ONE}, 14\penalty0 (9):\penalty0 e0221933, September 2019.
\newblock ISSN 1932-6203.
\newblock \doi{10.1371/journal.pone.0221933}.
\newblock URL \url{https://www.ncbi.nlm.nih.gov/pmc/articles/PMC6728051/}.

\bibitem[Stocking and Holstein(1993)]{stocking_constructing_1993}
S.~Holly Stocking and Lisa~W. Holstein.
\newblock Constructing and {Reconstructing} {Scientific} {Ignorance}:
  {Ignorance} {Claims} in {Science} and {Journalism}.
\newblock \emph{Knowledge}, 15\penalty0 (2):\penalty0 186--210, December 1993.
\newblock ISSN 0164-0259.
\newblock \doi{10.1177/107554709301500205}.
\newblock URL \url{https://doi.org/10.1177/107554709301500205}.
\newblock Publisher: SAGE Publications.

\bibitem[Ningrum and Atanassova(2023)]{Ningrum2023}
Panggih~Kusuma Ningrum and Iana Atanassova.
\newblock {Scientific Uncertainty: an Annotation Framework and Corpus Study in
  Different Disciplines}.
\newblock In \emph{19th International Conference of the International Society
  for Scientometrics and Informetrics (ISSI 2023)}, Bloomington, Indiana, US, 7
  2023.

\bibitem[Zimmermann(2000)]{zimmermann_application-oriented_2000}
H.~J. Zimmermann.
\newblock An application-oriented view of modeling uncertainty.
\newblock \emph{European Journal of Operational Research}, 122\penalty0
  (2):\penalty0 190--198, April 2000.
\newblock ISSN 0377-2217.
\newblock \doi{10.1016/S0377-2217(99)00228-3}.
\newblock URL
  \url{https://www.sciencedirect.com/science/article/pii/S0377221799002283}.

\bibitem[Powley and Dale(2007)]{powley_evidence-based_2007}
Brett Powley and Robert Dale.
\newblock \emph{Evidence-{Based} {Information} {Extraction} for {High}
  {Accuracy} {Citation} and {Author} {Name} {Identification}}.
\newblock January 2007.

\bibitem[Khedri and Kritsis(2020)]{khedri_how_2020}
Mohsen Khedri and Konstantinos Kritsis.
\newblock How do we make ourselves heard in the writing of a research article?
  {A} study of authorial references in four disciplines.
\newblock \emph{Australian Journal of Linguistics}, 40\penalty0 (2):\penalty0
  194--217, April 2020.
\newblock ISSN 0726-8602, 1469-2996.
\newblock \doi{10.1080/07268602.2020.1753011}.
\newblock URL
  \url{https://www.tandfonline.com/doi/full/10.1080/07268602.2020.1753011}.

\end{thebibliography}

\appendix
\renewcommand{\thefigure}{\thesection.\arabic{figure}} % Change to Appendix
\setcounter{figure}{0} % Reset figure numbering
\section{Appendix}
\label{Appendix:unscientify_demo}
% Set a margin for the specific column where the figures are placed
% \begin{adjustwidth}{-1cm}{0cm} 

\begin{figure}[b]
  \centering
  \includegraphics[width=0.9\columnwidth]{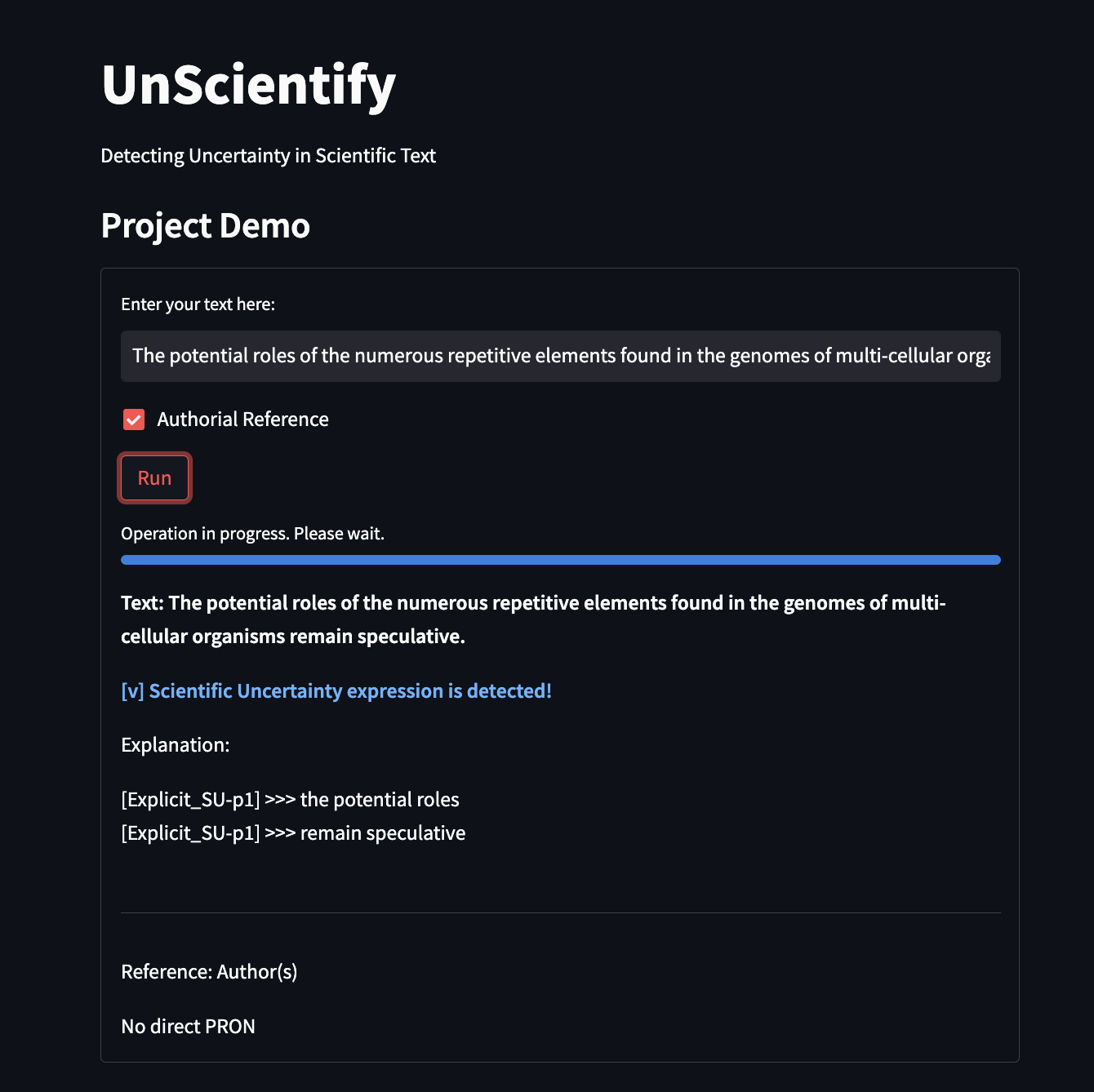}
  \caption{Demo 1 Detecting Explicit SU with multiple SU spans}
  \label{fig:demo1}
\end{figure}

\begin{figure}[b]
  \centering
  \includegraphics[width=1\columnwidth]{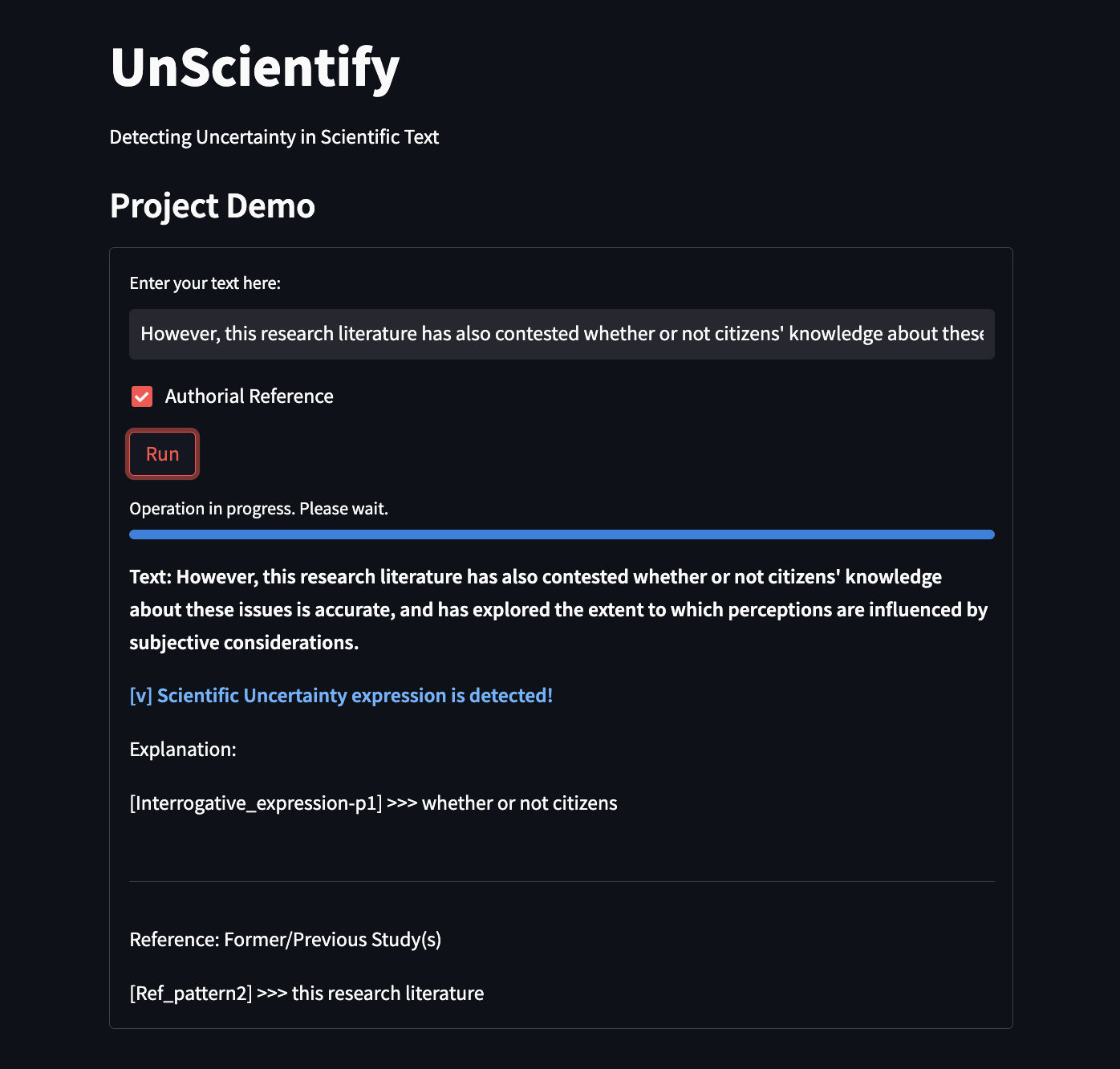}
  \caption{Demo 2 Detecting a sentence containing an interrogative expression}
  \label{fig:demo2}
\end{figure}

\begin{figure}[b]
  \centering
  \includegraphics[width=1\columnwidth]{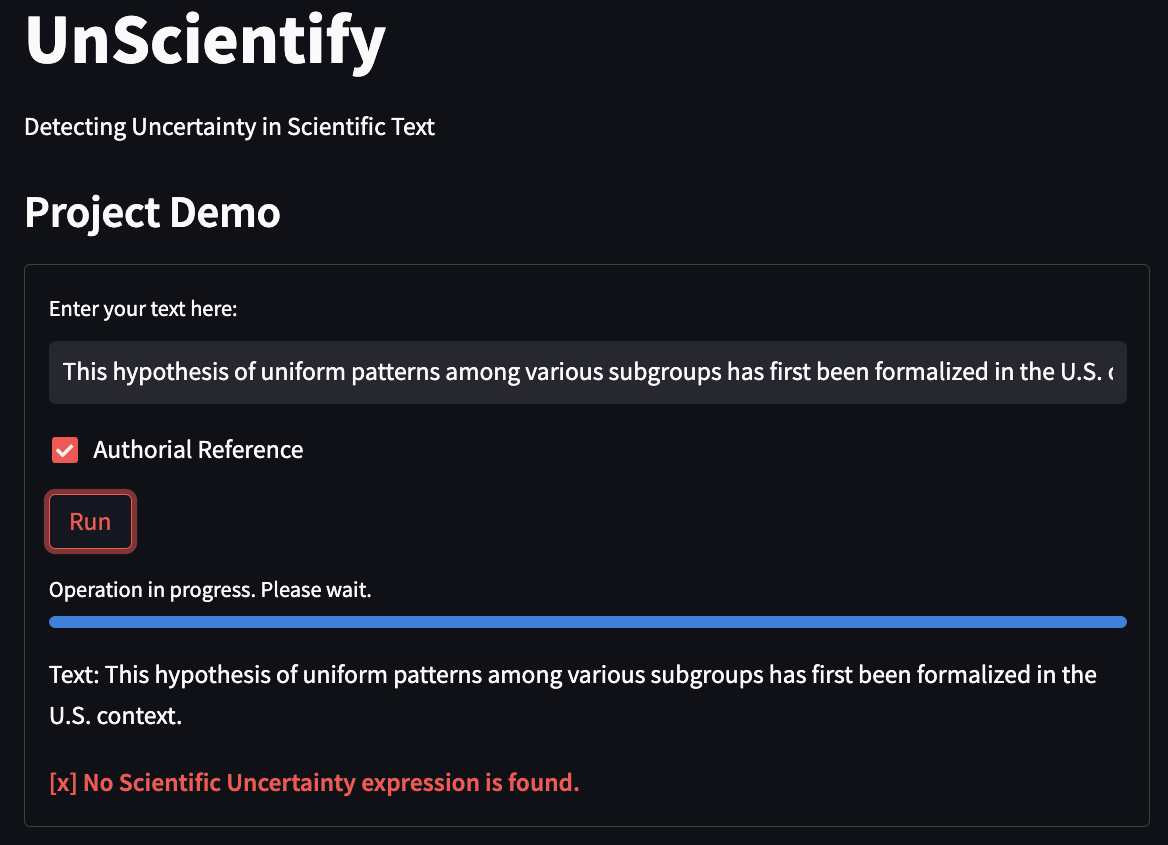}
  \caption{Demo 3 A sentence showing a Non-SU expression}
  \label{fig:demo3}
\end{figure}

\begin{figure}[b]
  \centering
  \includegraphics[width=1\columnwidth]{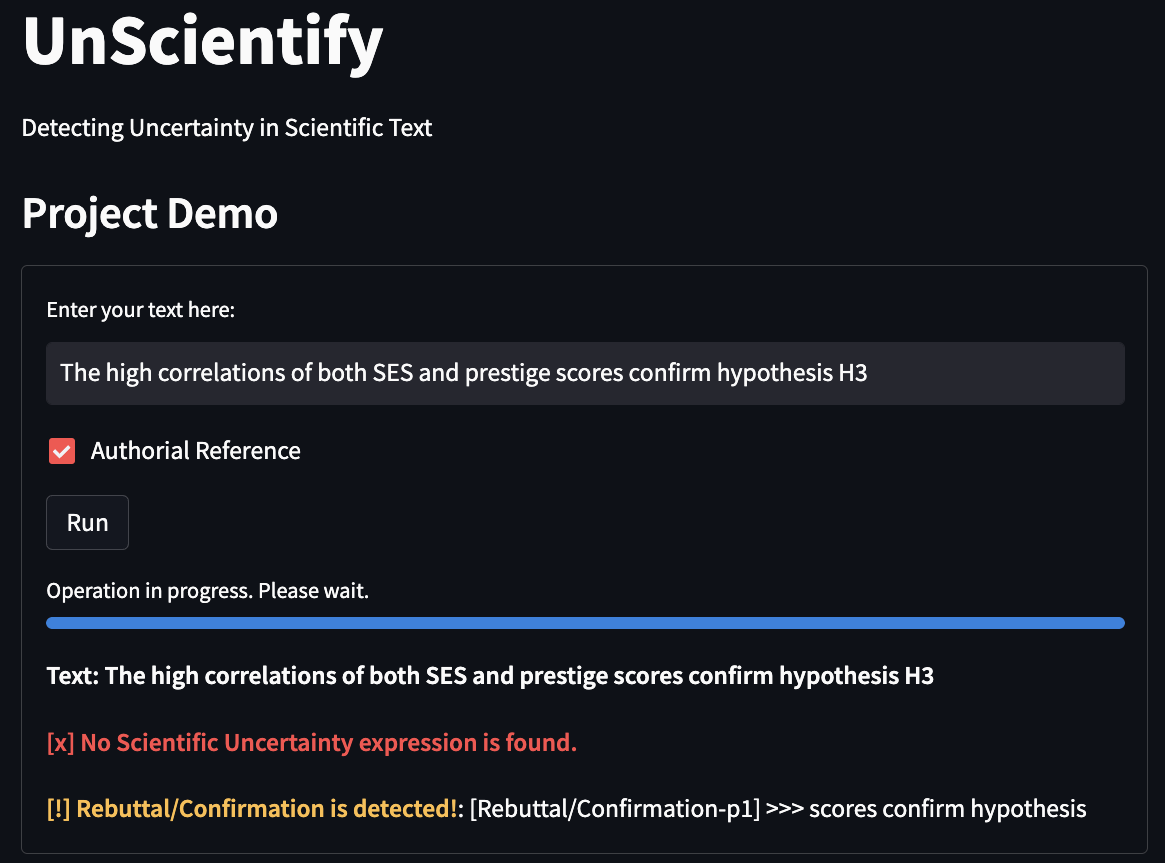}
  \caption{Demo 4 Rebuttal/Confirmation Statement Detection}
  \label{fig:demo4}
\end{figure}

% \begin{figure}[bt]
%   \centering
%   \begin{subfigure}{0.9\columnwidth}
%     \centering
%     \includegraphics[width=\textwidth]{Figures/explicit_SU.png}
%     \caption{Demo 1: Detecting Explicit SU with multiple SU spans}
%     \label{fig:demo1}
%   \end{subfigure}\hfill
%   \begin{subfigure}{0.9\columnwidth}
%     \centering
%     \includegraphics[width=\textwidth]{Figures/interrogative_SU.png}
%     \caption{Demo 2: Detecting a sentence containing an interrogative expression}
%     \label{fig:demo2}
%   \end{subfigure}\hfill
%   \begin{subfigure}{0.9\columnwidth}
%     \centering
%     \includegraphics[width=\textwidth]{Figures/hypothesis_Non.png}
%     \caption{Demo 3: A sentence showing a Non-SU expression}
%     \label{fig:demo3}
%   \end{subfigure}\hfill
%   \begin{subfigure}{0.9\columnwidth}
%     \centering
%     \includegraphics[width=\textwidth]{Figures/rebuttal.png}
%     \caption{Demo 4: Rebuttal/Confirmation Statement Detection}
%     \label{fig:demo4}
%   \end{subfigure}
%   \caption{UnScientify Demos}
%   \label{Appendix:unscientify_demo}
% \end{figure}

\end{document}